\documentclass{article}

\usepackage{microtype}
\usepackage{graphicx}
\usepackage{subfigure}
\usepackage{booktabs} \usepackage{hyperref}
\usepackage{enumitem}

\usepackage[accepted]{icml2025}

\usepackage{amsmath}
\usepackage{amssymb}
\usepackage{mathtools}
\usepackage{amsthm}

\usepackage[capitalize,noabbrev]{cleveref}

\theoremstyle{plain}

\theoremstyle{definition}

\theoremstyle{remark}

\usepackage[textsize=tiny]{todonotes}

\icmltitlerunning{Submission and Formatting Instructions for ICML 2025}

\usepackage{mathtools}
\everydisplay\expandafter{\the\everydisplay\vspace*{-0.5em}}

\begin{document}

\twocolumn[
    \icmltitle{Random Initialization Can’t Catch Up: The Advantage of Language Model Transfer for Time Series Forecasting}
    \icmltitlerunning{Random Initialization Can’t Catch Up: The Advantage of Language Model Transfer for Time Series Forecasting}

\icmlsetsymbol{equal}{*}
    
    \begin{icmlauthorlist}
    \icmlauthor{Roland Riachi}{equal,mila,umontreal}
    \icmlauthor{Kashif Rasul}{mstanley}
    \icmlauthor{Arjun Ashok}{mila,umontreal}
    \icmlauthor{Prateek Humane}{mila,umontreal}
    \icmlauthor{Alexis Roger}{mila,mcgill}\\
    \icmlauthor{Andrew R. Williams}{equal,mila,umontreal}
    \icmlauthor{Yuriy Nevmyvaka}{mstanley}
    \icmlauthor{Irina Rish}{mila,umontreal}
    \end{icmlauthorlist}
    
    \icmlaffiliation{umontreal}{University of Montreal}
    \icmlaffiliation{mila}{Mila - Québec AI Institute}
    \icmlaffiliation{mstanley}{Morgan Stanley}
    \icmlaffiliation{mcgill}{McGill University}
    
    \icmlcorrespondingauthor{Roland Riachi}{roland@riachi.com}
    \icmlcorrespondingauthor{Andrew R. Williams}{andrew.williams@umontreal.ca}
    
\icmlkeywords{Machine Learning, ICML}
    
    \vskip 0.3in
]

\printAffiliationsAndNotice{\icmlEqualContribution} 

\begin{abstract}
Recent works have demonstrated the effectiveness of adapting pre-trained language models (LMs) for forecasting time series in the low-data regime.
We build upon these findings by analyzing the effective transfer from language models to time series forecasting under various design choices including upstream post-training, time series tokenizer and language backbone size.
In the low-data regime, these design choices have a significant impact on the validation loss, with clear-cut choices that outperform others.
Contrary to \cite{hernandez2021scaling}, we observe that the validation loss of the LMs continues to smoothly decrease long after the validation loss of the randomly initialized models has converged, leading to a \textit{non-vanishing transfer gap} that holds across design choices.
These findings not only help shed light on the effective use of compute-efficient training for time series, but also open the way for the study of \textit{modality-agnostic} properties of data distributions leveraged by these models\footnote{Trained checkpoints and code will be released upon acceptance.}.
\vspace{-1em}
\end{abstract}

\section{Introduction}

Transfer learning enables the adaptation of models trained in one domain to downstream tasks in another. 
With the rise of large foundation models trained on massive text corpora, many specialized models have been created by fine-tuning these models for specific downstream applications \cite{bommasani2022opportunities}. Notably, it has been proven possible to achieve state-of-the-art results on downstream tasks significantly different from the original setting \cite{raffel2019t5}. 

\cite{lu2021pretrainedtransformersuniversalcomputation} explore this phenomenon through various synthetic and real world tasks. 
In this paper, we extend this line of work to univariate time series forecasting.
Similarly to text generation, time series forecasting is a sequence prediction problem, albeit with numerical values. This transition from discrete text tokens to continuous numerical values introduces new challenges in tokenization, embedding, and output representation.

Several recent studies have attempted to apply LLMs to time series forecasting using different techniques. Reprogramming-based methods, such as that of \citet{jin2023timellm}, train an encoder to align input summary statistics with text-like representations. Other works explore zero-shot forecasting by directly feeding digitized time series data into LLMs as natural language inputs \cite{gruver2024largelanguagemodelszeroshot,requeima2024llm,williams2024context}. Still others investigate parameter-efficient fine-tuning strategies. For instance, \citet{zhou2023fits} adapt normalization layers of GPT2, while \citet{ansari2024chronoslearninglanguagetime} train T5 models using bin-based tokenization. Their results show limited gains from pretrain initializing from pre-trained language models, in contrast to \citet{bayazigeneral}, who find positive transfer from T5 to EEG signals, which suggests that the properties of the time series data may have an impact on whether transfer occurs.

In this paper, we seek to contribute to this discussion by examining the transferability of language models to univariate time series forecasting. Specifically, we ask:

\begin{enumerate}[topsep=0pt, itemsep=3pt, parsep=0pt]
\item What is the most effective way to encode time series data into a latent space?
\item How does the upstream model’s performance influence downstream forecasting accuracy?
\end{enumerate}

To this end, we present the following contributions:
\begin{enumerate}[topsep=0pt, itemsep=3pt, parsep=0pt]
\item We demonstrate that transferring from pre-trained weights is more suitable in almost every scenario we test. Regardless of tokenization/embedding techniques or model size, models initialized with the weights of pre-trained language models outperform their randomly initialized counterparts.
    \item We quantify the amount of additional data samples necessary for a model trained from scratch to achieve the same loss on time series as a model initialized with pre-trained LLM weights.
In particular, our results differ from those of~\citep{hernandez2021scaling}, in that there exists a ``transfer gap'' such that no amount of additional tokens would enable a randomly initialized model to perform as well as the pretrained language initialization. 
\item We find that upstream model performance is not directly correlated with transfer to time series forecasting. Namely, while scaling parameter count increases effective transfer, upstream instruction tuning on language tasks results in worse downstream validation loss.

\end{enumerate}

\vspace{-0.3em}
\section{Experiments/Methodology}

To study the effective transfer of language to time series, we experiment with the T5\footnote{We use the efficient variants: https://huggingface.co/google/t5-efficient-base.} encoder-decoder transformer as the LLM backbone that we adapt to the time series modality. Our choice of T5 LLM is inspired by the initial success of \cite{bayazigeneral} with this model on EEG time series. The choice of LLM backbone is quite important as it may significantly affect the transfer, as we observed in our earlier experiments trying different language models; however, a systematic  ablation study with respect to the LLM backbone choice is out of scope of this paper.

The adaptation of LLM backbones to time series requires several architectural modifications, which we investigate.
First, we compare the performance of three different \textbf{weight initialization} strategies for the backbone.
Second, we replace the \textbf{input tokenization} and embedding layers to handle continuous-valued inputs.
Third, we vary the \textbf{backbone size}.

All models use the same output distribution as in \cite{ansari2024chronoslearninglanguagetime} and are trained using the LOTSA dataset (see Appendix A of \cite{woo2024unifiedtraininguniversaltime}) across 3 random seeds. For details regarding tokenizer design as well as initializations, see Appendix \ref{app:embeddings}.

\subsection{Effect of Backbone Initialization}

To assess the effective transfer from language weights to time series, we initialize the T5 architecture with three different sets of weights before fine-tuning with time series data:
\begin{enumerate}[topsep=0pt, itemsep=3pt, parsep=0pt]
    \item \textbf{Random initialization}: The baseline against which we compare transfer from language models is a random initialization according to Hugging Face defaults.
    \item \textbf{Purely pre-trained initialization}: We measure transfer from language by initializing the model with the weights of T5-Efficient~\cite{tay2022scaleefficientlyinsightspretraining}.
    \item \textbf{Instruction-tuned initialization}: We use the weights of Flan-T5 \citep{chung2022scalinginstructionfinetunedlanguagemodels}, an instruction-tuned version of T5.
\end{enumerate}

\subsection{Effect of Tokenizer} \label{sec:method:tokens}

A critical aspect of adapting transformer-based models for time series is the transformation of continuous time observations into tokens.
Given a time series sequence $\{x_t\}_{t=1}^T$ of length $T$, we aim to produce a sequence of tokens 
$\{s_t\}_{t=1}^{T'} \in \mathbb R^{d_{token} \times T'}$, where each token is of dimension $d_{token}$ and $T'\leq T$ denotes the length of the token sequence\footnote{Prior to tokenization, time series entries are normalized using the standard deviation across the input window to ensure consistent scales across different inputs. We consistently employ the same pre-processing steps to compare tokenization methodologies in isolation.
}.
We test three tokenization methods:
\begin{enumerate}[topsep=0pt, itemsep=3pt, parsep=0pt]
    \item \textbf{Naive tokenization}:
        The most straightforward approach; each time series observation is directly treated as a token, i.e., $s_t = x_t$.
    \item \textbf{Lag tokenization}:
        To incorporate seasonal dependencies, \citet{rasul2024lagllamafoundationmodelsprobabilistic} constructs token vectors using current and past time series values based on a set of lag indices $\mathcal L = \{\ell_1, \ldots, \ell_p\}$.
        Each token vector $s_t \in \mathbb R^{d_{token}}$ is given by
        $s_t = [x_t \,\, x_{t - \ell_1} \,\, \ldots \,\, x_{t - \ell_p}]^T.$
    \item \textbf{Bin tokenization}:
        To better align time series data with natural language tokens, \citet{ansari2024chronoslearninglanguagetime} bins time series values  into $B$ linearly-spaced bins within the range $[-a,a]$.
        Each token is assigned an index based on its bin.
\end{enumerate}

\subsection{Effect of Backbone Size}

\citet{lu2021pretrainedtransformersuniversalcomputation} find that performance scales reliably with model size for frozen pretrained language transformers adapted to CIFAR-10.
To investigate the effect of model size on transfer to time series, we test three sizes: \textbf{small} (60M), \textbf{base} (220M), and \textbf{large} (770M).
Intuitively, we hypothesize that using larger models leads to lower loss and more effective transfer in the low-data regime since larger models require more training to converge~\cite{Kaplan2020ScalingLF}; as a result, we expect effective transfer in the low-data regime to increase as model size increases.

\section{Results}
\vspace{-0.25em}
\subsection*{Do pre-trained weights lead to better performance than randomly initialized weights?}
\vspace{-0.25em}
\begin{figure}[!b]
    \vspace{-2em}
    \centering
    \includegraphics[width=\linewidth]{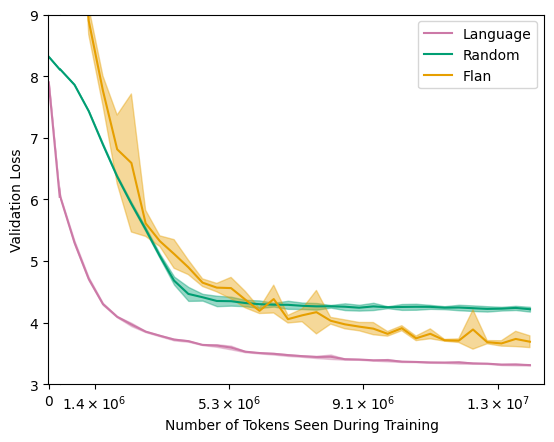}
    \vspace{-2em}
    \caption{
    \textbf{Validation losses for different weight initializations:} pre-training only, or ``Language'' (pink); pre-training + instruction-tuning, or ``Flan'' (yellow); and random initialization (green).
Both pre-training and instruction tuned weights quickly outperform the random initialization.
Furthermore, even though the Flan-T5 initialization outperforms solely pre-trained models on language tasks (see Table 5 of \citep{chung2022scalinginstructionfinetunedlanguagemodels}), Flan-T5 performs worse than the random initialization for the first 6M time series training tokens, and performs worse than the solely pre-trained initialization across the entire training duration.
    }
    \label{fig:lang_flan_rand_val_loss}
\end{figure}

Figure \ref{fig:lang_flan_rand_val_loss} demonstrates that language weights, both pre-training only (pink) and instruction-tuned (yellow), exhibit positive transfer against the random initialization (green).
In particular, at the end of training, both the pre-trained and instruction-tuned models exhibit a \textbf{non-vanishing transfer gap} that persists even as training concludes. While the performance of randomly initialized models plateaus, the performance of the language and instruction-tuned models continues to improve. This behavior contrasts with the findings of \citet{hernandez2021scaling}, where the transfer gap vanishes as the number of relevant training tokens increases.

However, in the low-data regime, the pre-trained weights and the instruction-tuned weights exhibit very different behaviours.
Whereas the pretrained model immediately and consistently outperforms the random initialization, the Flan-T5 initialization performs worse than the random initialization for the first 6M tokens.
As training progresses, the validation loss of the Flan-T5 initialization gradually converges towards that of the T5 model, although the curve qualitatively appears to be much noisier and spikier.

Simply choosing the best language model does not immediately return the best downstream performance for time series forecasting. 
Instead, these results suggest that post-training may have a complex impact on the representation space induced by the model's parameters, which improves language performance at the expense of a more jagged representation landscape that is less accommodating to transfer \cite{mosbach2021stabilityfinetuningbertmisconceptions}.

\subsection*{How do tokenization schemes affect effective-transfer?}
\begin{figure}[!b]
    \centering
    \begin{minipage}[b]{0.45\textwidth}
        \includegraphics[width=0.95\linewidth]{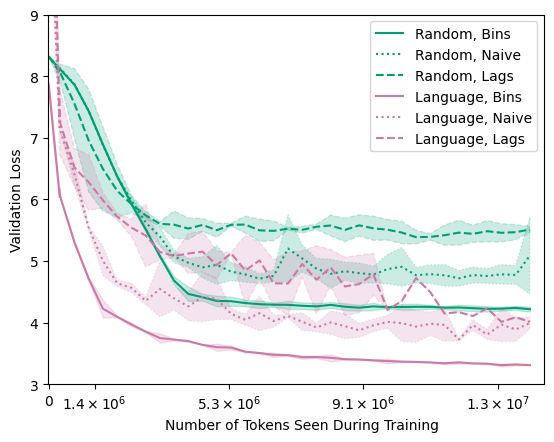}
\end{minipage}
    \hfill
    \begin{minipage}[b]{0.45\textwidth}
        \includegraphics[width=0.95\linewidth]{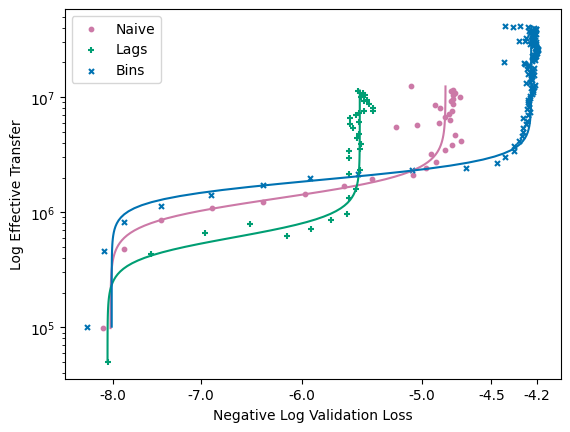}
\end{minipage}
    \vspace{-1em}
    \caption{
        (a) \textbf{Validation losses across tokenizers}. 
        Naive and lag tokenizations yield significantly worse zero-shot, i.e. initial, validation loss with pre-trained weights against random initializations.
        Nevertheless, after 1M time series training tokens, pre-trained models all have lower loses than their randomly initialized counterparts.
        Notably, each pre-trained model eventually achieves similar or lower validation loss compared to every randomly initialized model.
        (b)     \textbf{Effective transfer across tokenizers for T5 220M}.
        Defined as the difference in training tokens required to obtain a given validation loss (see Appendix \ref{app:effective_transfer}), effective transfer quantifies the additional number of training tokens necessary for the randomly initialized model to match the validation loss of the model initialized from pre-trained weights.
        Vertical asymptotes occur when the models with the randomly initialized weights converge before their pre-trained counterparts.
        We observe that log effective transfer increases with lower log validation loss, meaning that the models initialized from pre-trained weights converge model quickly than randomly initialized models.
    }
    \label{fig:lang_rand_tokenizer_combined}
\end{figure}

Figure \ref{fig:lang_rand_tokenizer_combined} (a) shows that transfer is beneficial regardless of tokenization/embedding scheme, but that the chosen tokenizer affects the validation loss and the convergence speed.

The bin tokenizer with pre-trained language weights (solid green line) achieves the smallest validation loss while converging smoothly and quickly.
We conjecture that this is because bin tokenization effectively shrinks the vocabulary and recasts forecasting as a classification task, which naturally aligns with the expectations of language models. 
This is supported by Chronos, which attributes its strong performance to the natural compatibility of quantization-based tokenization with language model architectures~\cite{ansari2024chronoslearninglanguagetime}.
This may also explain why the loss curves for the bin tokenizer are smoother than for the lag and linear tokenizers.

The lag tokenizer has the highest validation loss and is slow to converge, despite having access to the additional information provided by the lags.
This suggests that the model may be struggling to make use of this extra information, or that the information is not useful.

We also note that, across all three tokenizers, the pretrained initializations outperform their respective counterparts with random initializations.
Figure \ref{fig:lang_rand_tokenizer_combined} (a) clarifies that initializing the model from pre-trained language weights leads to meaningfully lower validation losses when compared with random weight initializations for all three tokenization methods, while figure (b) shows that this transfer is present in the low-data regime as well as at the end of training.

\subsection*{How does model size affect effective transfer?}

\begin{figure}[!t]
    \centering
    \includegraphics[width=\linewidth]{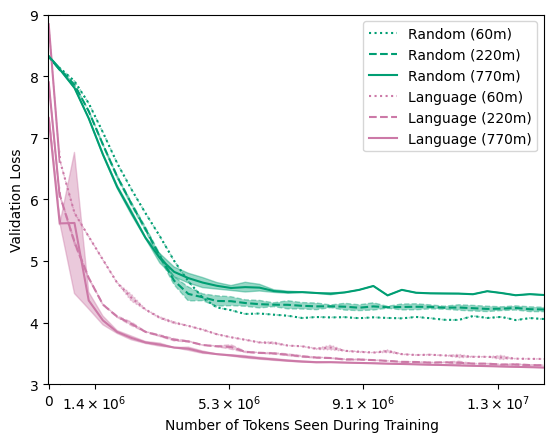}
    \vspace{-2em}
\caption{
    \textbf{Validation losses across backbones sizes.}
    Validation losses of models initialized with language weights decrease as model size increases, whereas those of randomly initialized do not.
    Moreover, across all model scales, the randomly initialized models converge early in training while
    the models with pre-trained weights did not converge within the length of training tested.
In the low-data regime, pretrained larger models outperform pretrained smaller models, but this difference tapers out as training continues.
    }
    \label{fig:scale_loss}
    \vspace{-2em}
\end{figure}

Consistent with~\citet{hernandez2021scaling}, we observe that increasing the number of model parameters influences the performance disparity between weight initializations.
\cref{fig:scale_loss} demonstrates that scaling model size with pre-trained initializations improves downstream performance on time series data in the low-data regime: the differences in validation loss between model sizes are most pronounced at the start of training, which aligns with the conjecture that larger models exhibit higher effective transfer in the low-data regime~\cite{hernandez2021scaling}.
In particular, models initialized with random weights exhibit an inverse relationship between scale and validation loss, highlighting the greater training needs of larger models starting from scratch.

We also note that the validation loss of the models with language initializations smoothly continues to improve with training, in contrast to the randomly initialized models which converge at a higher loss value and earlier.
This suggests that the language initialization meaningfully affects the loss landscape and training dynamics, resulting in asymptotically lower, predictably decreasing loss values on the downstream task of time series prediction.

\section{Conclusion and Future Work}

We find that training the T5 architecture on time series data performs better when initializing from pretrained language weights than from random weights.
These results hold across different experimental setups, including whether the language weights were post-trained or not, different choices of tokenizers and various model sizes.

We hypothesize that transfer is effective because pretraining biases the model toward a region of parameter space that is already well-suited for learning temporal patterns. 
While our experiments are limited to the T5 model family, we aim to test transfer across different language model backbones in future work to control for the impact of the architecture.

Another possibility is that time series data is distributionally similar to language data, and the benefit arises simply due to increased training data. To investigate this, we plan on controlling for total training data size in future work.

Interestingly, the Flan-T5 initialization underperforms against the T5 initialization. Therefore, further upstream tuning of language weights to improve language performance \citep{wei2022finetunedlanguagemodelszeroshot} does not translate to improvements in effective transfer to time series.
We plan to study this by ablating upstream datasets and fine-tuning approaches.

Finally, it is possible that the model learns to map time series data into a representation space similar to that of language. To investigate this, we plan to use statistical analysis tools, such as PQMass~\cite{lemos2024pqmass} to compare the distributions of time series and language representations before and after encoding. 

Ultimately, we hope to clarify whether transfer is effective because the models are trained with more data or because language and time series tasks share underlying structural similarity.

\newpage

\bibliography{ref}
\bibliographystyle{icml2025}

\newpage
\appendix
\onecolumn

\section{Additional Results and Visualizations}

\begin{figure}[h!]
    \centering

\end{figure}

\begin{figure}[h!]
    \centering
    \begin{minipage}[b]{0.45\textwidth}
        \includegraphics[width=1.1\linewidth]{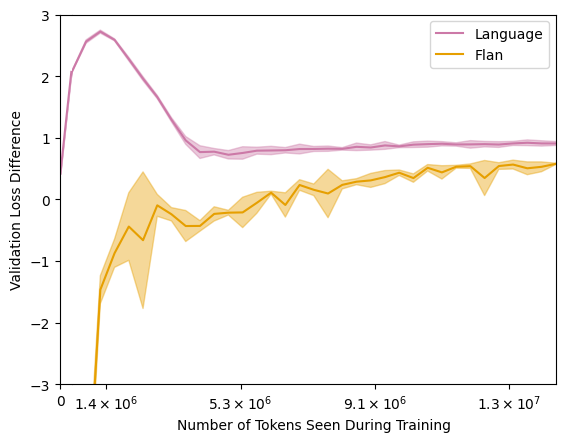}
        \caption{Difference in validation loss (pre-trained minus language) across weight initializations for the 220M parameter T5 backbone.}
        \label{fig:lang_flan_rand_loss_diff}
    \end{minipage}
    \hfill
    \begin{minipage}[b]{0.45\textwidth}
        \includegraphics[width=1.1\linewidth]{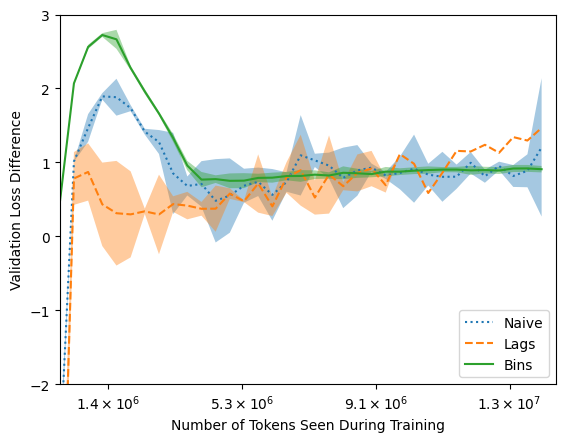}
        \caption{Difference in validation loss (random minus language) across tokenizers for the 220M parameter T5 backbone.}
        \label{fig:lang_rand_tokenizer_loss_diff}
    \end{minipage}
\end{figure}

\begin{figure}[h!]
    \centering
    \begin{minipage}[b]{0.45\textwidth}
        \includegraphics[width=1.1\linewidth]{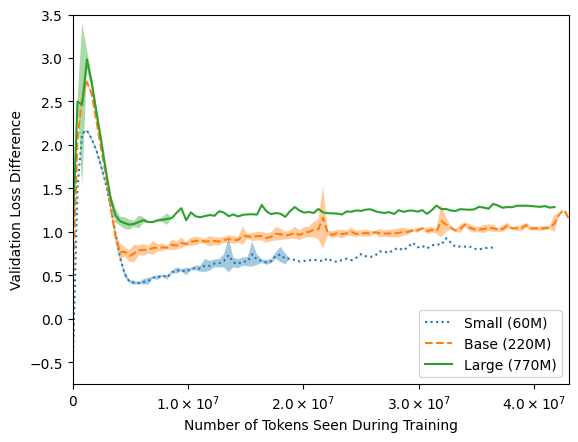}    
        \caption{Difference in validation loss (random minus language) across model backbone scales.}
        \label{fig:lang_rand_scale_val_loss_diff}
    \end{minipage}
    \hfill
    \begin{minipage}[b]{0.45\textwidth}
        \includegraphics[width=1.1\linewidth]{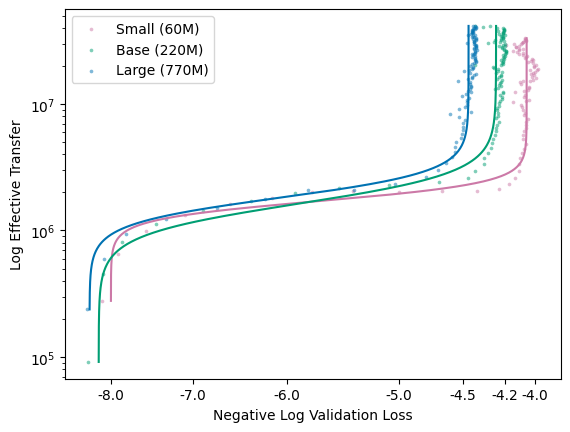}
        \caption{Effective transfer across model backbone scales.}
        \label{fig:lang_rand_scale_effective_transfer}
    \end{minipage}
\end{figure}

\section{Hyperparameter Search}

\begin{figure}[h!]
     \centering
     \includegraphics[width=0.5\linewidth]{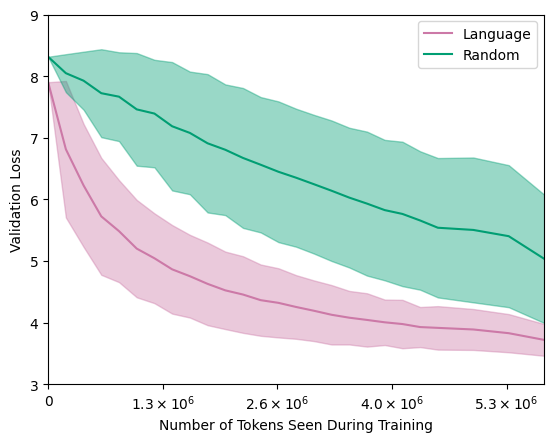}
      \caption{Mean validation loss across hyperparameter configurations.}
      \label{fig:hparam-agg}
 \end{figure}

\begin{figure}[h!]
     \centering
     \includegraphics[width=0.5\linewidth]{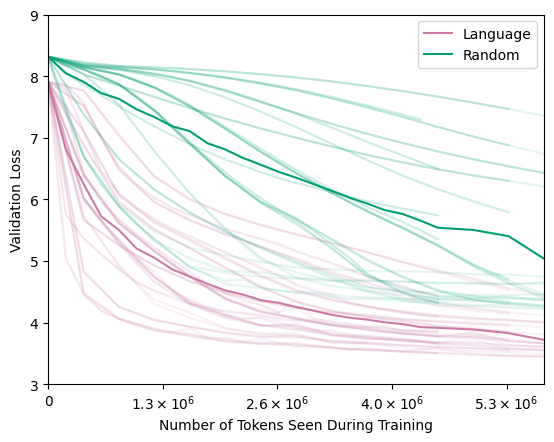}
     \caption{Validation losses across hyperparameter configurations.}
     \label{fig:hparam-samples}
 \end{figure}

When using a pre-trained model for downstream tasks, choices of hyperparameters are different compared to when training a model from scratch.
To avoid unfair comparisons caused by a given choice of hyperparameters, we perform a sweep across common settings for the 220M parameter T5 backbone with the bin tokenizer.
Figure \ref{fig:hparam-agg} shows the mean validation loss curve with standardard deviations across all such configurations.
Moreover, Figure \ref{fig:hparam-samples} shows the individual validation loss curves (transparent) across for all these configurations along with the mean validation loss curve for both language and random weights (opaque).
As can be seen from both figures, initializing the model with pre-trained weights consistently yields lower validation losses than random initializations, across most choices of hyperparameters.

We ablate across the following choices of hyperparameters:
\begin{itemize}
    \item learning rate: 0.0001, 0.0005, 0.001
    \item batch size: 64, 128
    \item weight decay: 0, 0.01, 0.1
    \item warm-up duration (as a percentage of maximum training duration): 0, 0.5\%, 1\%, 2\%
\end{itemize}

\section{Embeddings}\label{app:embeddings}

As mentioned above, language data is discretely valued whereas time series data is continuously valued.
This distinction affects how we usually embed tokens for LLMs, which explicitly assign learnable embedding vectors $e_t$ to each token $s_t$.

\subsection{Embeddings Architectures}

We employ different embedding methods for discrete and continuous time series tokenization methods:
\begin{enumerate}[topsep=0pt, itemsep=0pt, parsep=0pt]
    \item Discrete mappings (bins):
        We apply a standard embedding layer, where the embedding $e_t$ for token $s_t = i$ is a lookup from an embedding matrix based on the bin index $i$.
    \item Continuous mappings (naive, lags):
        We apply a linear transformation to the token vectors:
        \[
            e_t = W_es_t + b_e,
        \]
        where $W_e \in \mathbb R^{d_{model} \times d_{token}}$ and $b_e \in \mathbb R^{d_{model}}$.
\end{enumerate}

\subsection{Embeddings Pre-Trained Initializations}

For each tokenization approach, the embedding layers of the pre-trained model cannot be directly re-used, and instead require an intermediate processing step:
\begin{enumerate}[topsep=0pt, itemsep=0pt, parsep=0pt]
    \item Discrete mapping:
        Initialize the embedding matrix to the first $B=4096$ vocabulary vectors of the pre-trained backbone.
    \item Continuous mapping:
        Each row of the matrix $W_e$ is initialized to the mean vocabulary vector of the pre-trained model, while the bias $b_e$ is initialized as the zero vector.
\end{enumerate}

\section{Effective Transfer}\label{app:effective_transfer}

Let $\mathcal L_R(d)$ and $\mathcal L_P(d)$ denote the validation losses achieved after training for $d$ time series tokens using random and pre-trained initializations, respectively.
For a given validation loss $\ell$, $\mathcal L_\cdot^{-1}(\ell)$ therefore denotes the amount of data required to train a model to reach the loss level $\ell$ begining from a given initialization.
We define transfer as ``effective'' or ``positive'' when starting the timeseries model training with the language model's weights allows us to achieve the same validation loss with less data than a model initialised randomly. This data quantity difference is what we refer to as the amount of data we ``saved''. Concretely, the \textit{effective data transferred} $D_T(\ell)$ for a target loss $\ell$ is defined as the difference of these data amounts:
\[
    D_T(\ell) := \mathcal L_R^{-1}(\ell) - L_P^{-1}(\ell).
\]
A positive $D_T(\ell)$ indicates that initializing the model from pre-trained language weights requires fewer training examples to reach the loss $\ell$ compared to random initialization, signifying an effective transfer from the upstream task to time series forecasting.

\end{document}